\documentclass[conference]{IEEEtran}
\IEEEoverridecommandlockouts
\usepackage{cite}
\usepackage{amsmath,amssymb,amsfonts}
\usepackage{algorithmic}
\usepackage{graphicx}
\usepackage{textcomp}
\usepackage{tikz}
\usetikzlibrary{shapes}
\usepackage{xcolor}
\def\BibTeX{{\rm B\kern-.05em{\sc i\kern-.025em b}\kern-.08em
    T\kern-.1667em\lower.7ex\hbox{E}\kern-.125emX}}
    
\usepackage{eurosym}
\usepackage{booktabs}

\newcommand{\refequation}[1]{\eqref{#1}}
\newcommand{\reffigure}[1]{\figurename~\ref{#1}}
\newcommand{\refsection}[1]{Section~\ref{#1}}
\newcommand{\refsubsection}[1]{Subsection~\ref{#1}}
\newcommand{\reftable}[1]{Table~\ref{#1}}    

\newcommand{\transpose}[1]{#1^\top}

\begin{document}

\title{Learning Policies from Self-Play with\\ Policy Gradients and MCTS Value Estimates\\
\thanks{Funded by a \euro2m ERC Consolidator Grant (\texttt{http://ludeme.eu}).}}

\author{\IEEEauthorblockN{Dennis J. N. J. Soemers, {\'E}ric Piette, Matthew Stephenson, and Cameron Browne}
\IEEEauthorblockA{\textit{Department of Data Science and Knowledge Engineering} \\
\textit{Maastricht University}\\
Maastricht, the Netherlands \\
\texttt{\{dennis.soemers,eric.piette,matthew.stephenson,cameron.browne\}@maastrichtuniversity.nl}}
}

\maketitle

\begin{abstract}
In recent years, state-of-the-art game-playing agents often involve policies that are trained in self-playing processes where Monte Carlo tree search (MCTS) algorithms and trained policies iteratively improve each other. The strongest results have been obtained when policies are trained to mimic the search behaviour of MCTS by minimising a cross-entropy loss. Because MCTS, by design, includes an element of exploration, policies trained in this manner are also likely to exhibit a similar extent of exploration. In this paper, we are interested in learning policies for a project with future goals including the extraction of interpretable strategies, rather than state-of-the-art game-playing performance. For these goals, we argue that such an extent of exploration is undesirable, and we propose a novel objective function for training policies that are not exploratory. We derive a policy gradient expression for maximising this objective function, which can be estimated using MCTS value estimates, rather than MCTS visit counts. We empirically evaluate various properties of resulting policies, in a variety of board games.
\end{abstract}


\section{Introduction}

Monte Carlo tree search (MCTS) algorithms \cite{Kocsis2006UCT, Coulom2007}, often in combination with learning algorithms, provide state-of-the-art AI in many games and other domains \cite{Browne2012, Silver2017AlphaGoZero, Anthony2017ExIt, Silver2018AlphaZero}. The most straightforward implementations of MCTS use large numbers of play-outs where actions are selected uniformly at random to estimate the value of the starting state of those play-outs. Play-outs using handcrafted heuristics, learned policies, or search to more closely resemble realistic lines of play can often significantly increase playing strength, even if the increased computational cost leads to a reduction in the number of play-outs \cite{Gelly2006ModificationUCT, Coulom2007EloRatingsPatterns, Gelly2007CombiningOnlineOffline, Silver2009MCSimulationBalancing, Baier2010PowerForgetting, Huang2010MCSimulationBalancingPractice, Winands2011AlphaBetaPlayouts, Nijssen2012PlayoutSearch, Silver2012TDSearch, Graf2016AdaptivePlayouts, Graf2016MCSimulationBalancingRevisited, SilverHuangEtAl16nature, Cazenave16PlayoutPolicyAdaptation, Anthony2017ExIt}.

The majority of policy learning approaches use supervised learning with human expert moves as training targets, or traditional reinforcement learning (RL) update rules \cite{SuttonBarto2018RLBook}, but the most impressive results have been obtained using the Expert Iteration framework, where MCTS and a learned policy iteratively improve each other through self-play \cite{Silver2017AlphaGoZero, Anthony2017ExIt, Silver2018AlphaZero}. In this framework, a policy is trained to mimic the MCTS search behaviour using a cross-entropy loss, and the policy is used to bias the MCTS search. Note that play-outs are sometimes replaced altogether by trained value function estimators, leaving only the selection phase of MCTS to be biased by a trained policy \cite{Silver2017AlphaGoZero, Silver2018AlphaZero}, but a learned policy may also be used to run play-outs \cite{Anthony2017ExIt}.

The selection phase of MCTS provides a balance between \textit{exploration} and \textit{exploitation}; exploration consists of searching parts of the game tree that have not yet been thoroughly searched, and exploitation consists of searching parts of the game tree that appear the most promising based on the search process so far. Using the search behaviour of MCTS as an update target for a policy means that this policy is trained to have a similar balance between exploration and exploitation as the MCTS algorithm.

Within the context of the Digital Ludeme Project \cite{Browne2018ModernTechniques}, we aim to learn policies based on interpretable features \cite{Browne2019StrategicFeatures} for state-action pairs, where future goals of the project include extracting explainable strategies from learned policies, and estimating similarities or distances between different (variants of) games in terms of strategies. For the purpose of these goals, we do not expect the exploratory behaviour that is learned with the standard cross-entropy loss to be desirable.

We formulate a new training objective for policies. A policy that optimises this objective can intuitively be understood as one that selects actions such that MCTS is subsequently expected to be capable of performing well. Unlike the case where the MCTS search behaviour is used as training target, this optimisation criterion does not encourage any level of exploration. We derive an expression for the gradient of this objective with respect to a differentiable policy's parameters, which allows for training using gradient descent. 

Like the standard updates used to optimise the cross-entropy loss in Expert Iteration \cite{Silver2017AlphaGoZero, Anthony2017ExIt, Silver2018AlphaZero}, these updates are guided by ``advice'' generated by MCTS. This is hypothesized to be important for a stable and robust self-play learning process, with a reduced risk of overfitting to the self-play opponent. The primary difference is that this advice consists of value estimates, rather than a distribution over actions.

We empirically compare policies trained to optimise the proposed objective function, with policies trained on the standard cross-entropy loss, across a variety of deterministic, perfect-information, two-player board games. The proposed objective consistently leads to policies that are at least as strong, and in some games significantly stronger, than the cross-entropy loss. We also confirm that the resulting policies lead to significantly lower entropy in distributions over actions, which suggests that learned policies are less exploratory. Finally, we compare the resulting distributions of weights learned for different features, and the performance of MCTS agents biased by policies trained on the different objectives.

\section{Background}

This section formalises the concepts from reinforcement learning (RL) theory required in this paper. We assume a standard single-agent setting. When subsequently applying these concepts to multi-player, adversarial game settings, any states in which a learning agent is not the player to move are ignored, and moves selected by opponents are simply assumed to be a part of the ``environment'' and its transition dynamics.

\subsection{Markov Decision Processes}

We use the standard single-agent, fully-observable, episodic Markov decision process (MDP) setting, where $\mathcal{S}$ denotes a set of states, and $\mathcal{A}$ denotes a set of actions. At discrete time steps $t = 0, 1, \dots$, the agent observes states $S_t \in \mathcal{S}$. Whenever $S_t$ is not terminal, the agent selects an action $A_t \in \mathcal{A}(S_t)$ from the set of actions $\mathcal{A}(S_t)$ that are legal in $S_t$, which leads to an observed reward $R_{t+1} \in \mathbb{R}$. We assume that there is a fixed starting state $s_0$. Given a current state $s$ and action $a$, the probability of observing any arbitrary successor state $s'$ and reward $r$ is given by $p(s', r \mid s, a) = \text{Pr\{} S_t = s', R_t = r \mid S_{t - 1} = s, A_{t-1} = a \text{\}}$.

Let $\pi$ denote some policy, such that $\pi(s, a)$ denotes the probability of selecting an action $a$ in a state $s$, and $\sum_{a \in \mathcal{A}(s)} \pi(s, a) = 1$. The value $V^{\pi}(s)$ of a state $s$ under policy $\pi$ is given by \refequation{Eq:StateValue}:
\begin{equation} 
\label{Eq:StateValue}
    V^{\pi}(s) \doteq \mathbb{E} \left[ \sum_{t=0}^{\infty} \gamma^t R_{t+1} \mid S_0 = s, A_t \sim \pi \right],
\end{equation}
where $0 \leq \gamma \leq 1$ denotes a discount factor (in the board games applications considered in this paper, typically $\gamma = 1$). We define $R_t \doteq 0$ for $t > T$ in any episode where $S_T$ is a terminal state. The value $Q^{\pi}(s, a)$ of an action $a$ in a state $s$ under policy $\pi$ is given by \refequation{Eq:StateActionValue}:
\begin{equation} 
\label{Eq:StateActionValue}
    Q^{\pi}(s, a) \doteq \mathbb{E} \left[ \sum_{t=0}^{\infty} \gamma^t R_{t+1} \mid S_0 = s, A_0 = a, A_{>0} \sim \pi \right],
\end{equation}
where $A_{>0}$ covers all actions $A_t$ where $t > 0$.

\subsection{Policy Gradients}

Let $J(\pi)$ denote the expected performance, in terms of returns per episode, of a policy $\pi$:
\begin{equation}
\label{Eq:PolicyExpectedReturns}
J(\pi) \doteq \mathbb{E} \left[ \sum_{t=0}^{\infty} \gamma^t R_{t+1} \mid S_0 = s_0, A_t \sim \pi \right] = V^{\pi}(s_0).
\end{equation}
A common goal in RL is to find a policy $\pi$ such that this objective is maximised. Suppose that $\pi(\cdot, \cdot)$ is a differentiable function, parameterised by a vector $\boldsymbol{\theta}$, such that $\nabla_{\boldsymbol{\theta}} \pi(\cdot, \cdot)$ exists. Then, the Policy Gradient Theorem \cite{Sutton2000PolicyGradient} states that:
\begin{equation}
\label{Eq:PolicyGradientTheorem}
\nabla_{\boldsymbol{\theta}} J(\pi) = \sum_{s \in \mathcal{S}} d^{\pi}(s) \sum_{a \in \mathcal{A}(s)} \nabla_{\boldsymbol{\theta}} \pi(s, a) Q^{\pi}(s, a),
\end{equation}
where $d^{\pi}(s) \doteq \sum_{t=0}^{\infty} \gamma^t \text{Pr\{} S_t = s \mid S_0 = s_0, A_{<t} \sim \pi \text{\}}$ gives a discounted weighting of states according to how likely they are to be reached in trajectories following $\pi$. Sample-based estimators of this gradient allow for the objective to be optimised directly, using stochastic gradient ascent to adjust the policy parameters $\boldsymbol{\theta}$ \cite{Williams1992REINFORCE, Schulman2016GAE, SuttonBarto2018RLBook}.

\subsection{Monte Carlo Tree Search Value Estimates}

Most variants of Monte Carlo tree search (MCTS) \cite{Browne2012} can be viewed as RL approaches which, based on simulated experience, learn on-policy value estimates for the states represented by nodes in the search tree that is gradually built up \cite{Vodopivec2017MCTSRL}. Let $\sigma$ denote a state from which we run an MCTS search process (meaning that $\sigma$ corresponds to the root node). Then we can formally describe a policy $\mathcal{M}_{\sigma}$:
\begin{equation}
\label{Eq:MCTSPolicy}
\mathcal{M}_{\sigma}(s, a) =
\begin{cases}
\frac{N(s, a)}{\sum_{a'} N(s, a')} & \text{if } $s$ \text{ in search tree,} \\
\rho(s, a) & \text{otherwise,}
\end{cases}
\end{equation}
where $N(s, a)$ denotes the number of times that the search process selected $a$ in the node representing $s$, and $\rho(s, a)$ denotes the roll-out policy.

Suppose that value estimates $\hat{V}(s)$ in nodes of the search tree are computed, as is customary, as the averages of backpropagated scores, or using some other approach that can be viewed as implementing on-policy backups -- such as Sarsa-UCT($\lambda$) \cite{Vodopivec2017MCTSRL}. These value estimates are then unbiased estimators of $V^{\mathcal{M}_{s}}$, as defined in \refequation{Eq:StateValue}. We typically expect these value estimates to be unreliable and exhibit high variance deep in the search tree, but, given a sufficiently high MCTS iteration count, they may be more reliable close to the root node.

\section{Policy Gradient with MCTS Value Estimates} \label{Sec:PolicyGradientsMCTSValues}

Unlike the standard cross-entropy loss used in Expert Iteration, optimising the policy gradient objective of \refequation{Eq:PolicyExpectedReturns} does not incentivise an element of exploration in trained policies. However, this objective focuses on the long-term performance of the standalone policy $\pi$ being trained. 
Suppose that it is infeasible to learn a good distribution $\pi(s, \cdot)$ over actions in some state $s$ -- for instance because there are no features available that allow distinguishing between any actions in $s$. Reaching $s$ will then be detrimental to the long-term performance of $\pi$ according to \refequation{Eq:PolicyExpectedReturns}, and actions leading to $s$ will therefore be disincentivized, even if they may otherwise clearly be a part of the principal variation. This is problematic when we aim to use $\pi$ for purposes such as strategy extraction (even if only for some parts of the state space), rather than using it for standalone game-playing.


\subsection{Objective Function}

To address the issues illustrated above, we propose to maximise the objective function given by \refequation{Eq:NewObjective}, where $\pi$ is the apprentice policy to be trained, parameterised by a vector $\boldsymbol{\theta}$:
\begin{equation}
\label{Eq:NewObjective}
J_{TSPG}(\pi) \doteq \sum_{t=0}^{\infty} \mathbb{E} \left[ \gamma^t R_{t+1} \mid S_0 = s_0, A_t \sim \pi, A_{\neq t} \sim \mathcal{M} \right],
\end{equation}
where $A_{\neq t} \sim \mathcal{M}$ denotes that, for all $t' \neq t$, we run an MCTS process $\mathcal{M}_{S_{t'}}$ and sample $A_{t'}$ from $\mathcal{M}_{S_{t'}}(S_{t'}, \cdot)$. We refer to this as the \textit{Tree-Search Policy Gradient} (TSPG) objective function. Intuitively, sampling actions $A_{t'}$ for $t' < t$ from MCTS can be understood as stating that it is only important for $\pi$ to be well-trained in states that are likely to be reached when playing according to MCTS processes prior to time $t$. Sampling actions $A_{t'}$ for $t' > t$ from MCTS in this objective can be understood as stating that $\pi$ is not required to be capable of playing well for the remainder of an episode, but only needs to be able to select actions such that MCTS would be expected to perform well in subsequent states.

Suppose that there is a small game tree, in which MCTS can easily find an optimal line of play, but where that optimal line of play leads to a subtree in which a parameterised policy $\pi$ cannot play well. This may, for instance, be due to a lack of representational capacity of $\pi$ itself (i.e. using a simple linear function), or due to using a restricted set of input features that is insufficient for states or actions in that subtree to be distinguished from each other. A standard RL objective function, such as the one in \refequation{Eq:PolicyExpectedReturns}, would lead to a policy that learns to avoid that subtree altogether, because the same policy cannot guarantee long-term success in that subtree. We argue that this is detrimental for our goal of interpretable strategy extraction, because it leads to a poor strategy in the root of such a game tree. In contrast, the TSPG objective still allows for a strong strategy to be learned for states other than those in the problematic subtree.

\subsection{Policy Gradient} \label{SubSec:PolicyGradientDerivation}

Our derivation of an expression for the gradient of this objective with respect to the parameters $\boldsymbol{\theta}$ takes inspiration from the original proof for the policy gradient theorem \cite{Sutton2000PolicyGradient}. We start by defining $V^{\pi \mathcal{M}}(s)$ as the expected value of sampling a single action from $\pi$ in state $s$, and sampling actions from MCTS search processes for the remainder of the episode:
\begin{equation}
\begin{aligned}
V^{\pi \mathcal{M}}(s) &\doteq \mathbb{E} \left[ \sum_{t=0}^{\infty} \gamma^t R_{t+1} \mid S_0 = s, A_0 \sim \pi, A_{>0} \sim \mathcal{M} \right] \\
&= \sum_{a} \pi(s, a) Q^{\mathcal{M}}(s, a),
\end{aligned}
\end{equation}
where $\mathcal{M}$ is used as a shorthand notation to indicate that a separate policy $\mathcal{M}_{S_t}$, involving a separate complete search process, is used at every time $t$. The gradient of this function with respect to $\boldsymbol{\theta}$ is given by:
\begin{equation}
\label{Eq:GradientValueFunction}
\begin{aligned}
\nabla_{\boldsymbol{\theta}} V^{\pi \mathcal{M}}(s) &= \nabla_{\boldsymbol{\theta}} \sum_{a} \pi(s, a) Q^{\mathcal{M}}(s, a) \\
&= \sum_a \Big[ \nabla_{\boldsymbol{\theta}} \pi(s, a) Q^{\mathcal{M}}(s, a)\\
&\qquad \quad + \pi(s, a) \nabla_{\boldsymbol{\theta}} Q^{\mathcal{M}}(s, a) \Big] \\
&\approx \sum_a \left[ \nabla_{\boldsymbol{\theta}} \pi(s, a) Q^{\mathcal{M}}(s, a) \right],
\end{aligned}
\end{equation}
where we assume that $\nabla_{\boldsymbol{\theta}} Q^{\mathcal{M}}(s, \cdot) = 0$. 
Note that this assumption may be violated in practice by making use of $\boldsymbol{\theta}$ in the play-outs of MCTS processes, but it is not feasible to accurately estimate the gradient of the performance of MCTS with respect to parameters $\boldsymbol{\theta}$ used in play-outs. We can avoid violating the assumption by freezing the versions of parameters used for biasing any MCTS process, and clearing any old experience when updating parameters used by MCTS, but in practice we expect this to be detrimental to learning speed. Also note that this assumption is very similar to the omission of the $\pi(s, a) \nabla_{\mathbf{u}} Q^{\pi, \gamma}(s, a)$ term in the Off-Policy Policy-Gradient Theorem, where $\mathbf{u}$ is a parameter vector and $\pi$ is a target policy \cite{Degris2012OffPAC}.

Now, we rewrite the TSPG objective function to a more convenient expression, starting from \refequation{Eq:NewObjective}:
\begin{equation}
\begin{aligned}
J_{TSPG}(\pi) &\doteq \sum_{t=0}^{\infty} \mathbb{E} \left[ \gamma^t R_{t+1} \mid S_0 = s_0, A_t \sim \pi, A_{\neq t} \sim \mathcal{M} \right] \\
&= \sum_{s \in \mathcal{S}} d^{\mathcal{M}}(s) V^{\pi \mathcal{M}}(s),
\end{aligned}
\end{equation}
where $d^{\mathcal{M}}(s) = \sum_{t=0}^{\infty} \gamma^t \text{Pr\{} S_t = s \mid S_0 = s_0, A_{<t} \sim \mathcal{M} \text{\}}$. Taking the gradient with respect to $\boldsymbol{\theta}$ gives:
\begin{equation}
\label{Eq:AnalyticalExpressionGradient}
\begin{aligned}
\nabla_{\boldsymbol{\theta}} J_{TSPG}&(\pi) = \nabla_{\boldsymbol{\theta}} \sum_{s \in \mathcal{S}} d^{\mathcal{M}}(s) V^{\pi \mathcal{M}}(s) \\
&= \sum_{s \in \mathcal{S}} \left[ \nabla_{\boldsymbol{\theta}} d^{\mathcal{M}}(s) V^{\pi \mathcal{M}}(s) + d^{\mathcal{M}}(s) \nabla_{\boldsymbol{\theta}} V^{\pi \mathcal{M}}(s) \right] \\
&\approx \sum_{s \in \mathcal{S}} d^{\mathcal{M}}(s) \sum_a \nabla_{\boldsymbol{\theta}} \pi(s, a) Q^{\mathcal{M}}(s, a),
\end{aligned}
\end{equation}
where again we assume that $\boldsymbol{\theta}$ has no effect on MCTS processes by taking $\nabla_{\boldsymbol{\theta}} d^{\mathcal{M}}(\cdot) = 0$. 

The analytical expression of the gradient of the TSPG objective in \refequation{Eq:AnalyticalExpressionGradient} is exact if the involved MCTS processes are unaffected by $\boldsymbol{\theta}$, or an approximation otherwise. Note that it has a similar form to the original policy gradient expression in \refequation{Eq:PolicyGradientTheorem}. The weighting of states and the value estimates are now both provided by $\mathcal{M}$, but the only required gradient is for $\pi(\cdot, \cdot)$ (which, by assumption, is differentiable).

\subsection{Estimating the Gradient}

In the Expert Iteration framework \cite{Silver2017AlphaGoZero, Anthony2017ExIt, Silver2018AlphaZero}, experience is typically generated by playing self-play games where actions are selected proportional to the visit counts in root states after running MCTS processes. This corresponds precisely to the definition of policies $\mathcal{M}$ given in \refequation{Eq:MCTSPolicy}. It is customary to store states encountered in such a self-play process in a dataset $\mathcal{D}$ -- keeping only one randomly-selected state per full game, to avoid excessive correlations between instances -- and sample batches from $\mathcal{D}$ for stochastic gradient descent updates. Sampling batches of states $B \subseteq \mathcal{D}$ leads to unbiased estimates $\hat{g}$ of the gradient expression in \refequation{Eq:AnalyticalExpressionGradient}:
\begin{equation}
\label{Eq:GradientEstimate}
\hat{g} = \frac{1}{\vert B \vert} \sum_{s \in B} \left[ \sum_{a \in \mathcal{A}(s)} \nabla_{\boldsymbol{\theta}} \pi(s, a) Q^{\mathcal{M}}(s, a) \right].
\end{equation}

Optimisation of the cross-entropy loss typically used in Expert Iteration requires storing MCTS visit counts $N(s, a)$ for all $a \in \mathcal{A}(s)$ in the dataset $\mathcal{D}$, alongside the states $s$. Instead of storing visit counts, our approach requires storing MCTS value estimates $\hat{Q}(s, a)$ for all actions $a$ -- these are simply the state-value estimates $\hat{V}(s')$ of all successors $s'$ of $s$. These values can be plugged into \refequation{Eq:GradientEstimate} as unbiased estimators for $Q^{\mathcal{M}}(s, a)$.

We now have an unbiased estimator of the gradient which can be readily computed from data collected as in the standard Expert Iteration self-play framework. The form of this estimator most closely resembles that of the Mean Actor-Critic \cite{Allen2018MeanActorCritic}, in the sense that we explicitly sum over all actions rather than sampling trajectories with actions selected according to $\pi$. As in the gradient estimator of the Mean Actor-Critic, it is unnecessary to subtract a state-dependent baseline from $Q^{\mathcal{M}}(s, a)$ for variance reduction, as is typically done in sample-based estimators of policy gradients \cite{Sutton2000PolicyGradient, Schulman2016GAE}.

\section{Learning Offsets from Exploratory Policy} \label{Sec:LearningOffsets}

A differentiable policy $\pi$ is typically implemented to compute logits $z(s, a) = \transpose{\boldsymbol{\theta}} \boldsymbol{\phi}(s, a)$, where $\boldsymbol{\theta}$ is a trainable parameter vector and $\boldsymbol{\phi}(s, a)$ is a feature vector for a state-action pair $(s, a)$. Probabilities $\pi(s, a)$ are subsequently computed using the softmax function; $\pi(s, a) = \frac{\exp(z(s, a))}{\sum_{a'} \exp(z(s, a'))}$. In preliminary testing, we found that there is a risk for strong features that are only discovered and added in the middle of a self-play training process \cite{Soemers2019BiasingMCTS} to remain unused. When this happens, it appears like the learning approach remains stuck in what used to be a local optimum given an older feature set, even though newly-added features should enable escaping that local optimum. First, we elaborate on why this can happen, and subsequently propose an approach to address this issue.

\subsection{Gradients for Low-probability Actions}

Suppose that $\pi$ uses the softmax function, as described above. Then, the gradient of $\pi(s, a)$ with respect to the $i^{th}$ parameter $\theta_i$ of the parameter vector $\boldsymbol{\theta}$ is given by
\begin{equation}
\label{Eq:SoftmaxGradientSingleParam}
\nabla_{\theta_{i}} \pi(s, a) = \pi(s, a) \sum_{a'} \left( \delta_{aa'} - \pi(s, a') \right) \phi_i(s, a'),
\end{equation}
where the Kronecker delta $\delta_{aa'}$ is equal to $1$ if $a = a'$, or $0$ otherwise, and $\phi_i(s, a')$ denotes the $i^{th}$ feature value for the state-action pair $(s, a')$. 

This is the gradient that is multiplied by $Q^{\mathcal{M}}(s, a)$ in \refequation{Eq:GradientEstimate} to compute the update for the parameter $\theta_i$ corresponding to the feature $\phi_i$. In cases where features value $\phi_i(s, a)$ correlate strongly with state-action values $Q^{\mathcal{M}}(s, a)$, we would intuitively expect to obtain consistent, high-value gradient estimates to rapidly adapt $\theta_i$. However, if previous learning steps -- possibly taken before the feature $\phi_i$ was being used at all -- resulted in a parameter vector $\boldsymbol{\theta}$ such that $\pi(s, a)$ is low (i.e., $\pi(s, a) \approx 0$), this gradient will also be close to zero and learning progresses very slowly.

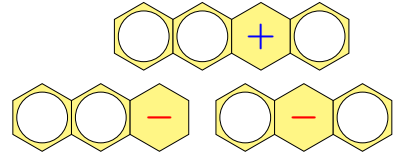
\begin{figure}[t]
\centering
\scalebox{0.9}{
    \begin{tikzpicture} [hexa/.style= {shape=regular polygon,regular polygon sides=6,minimum size=1cm,draw,inner sep=0,fill=white!40!yellow,shape border rotate=30},
    whitepiece/.style= {shape=circle,minimum size=0.75cm,draw,inner sep=0,fill=white}]
    
    \node at (0,0) [hexa] () {};
    \node at ({sin(60)}, 0) [hexa] () {};
    \node at ({2*sin(60)}, 0) [hexa] () {\Large $\color{blue}{\boldsymbol{+}}$};
    \node at ({3*sin(60)}, 0) [hexa] () {};
    
    \node at (0,0) [whitepiece] () {};
    \node at ({sin(60)},0) [whitepiece] () {};
    \node at ({3*sin(60)},0) [whitepiece] () {};
    
    \node at (-1.5,-1.2) [hexa] () {};
    \node at ({-1.5 + sin(60)}, -1.2) [hexa] () {};
    \node at ({-1.5 + 2*sin(60)}, -1.2) [hexa] () {\Large $\color{red}{\boldsymbol{-}}$};
    
    \node at (-1.5,-1.2) [whitepiece] () {};
    \node at ({-1.5 + sin(60)},-1.2) [whitepiece] () {};
    
    \node at (1.5,-1.2) [hexa] () {};
    \node at ({1.5 + sin(60)}, -1.2) [hexa] () {\Large $\color{red}{\boldsymbol{-}}$};
    \node at ({1.5 + 2*sin(60)}, -1.2) [hexa] () {};
    
    \node at (1.5,-1.2) [whitepiece] () {};
    \node at ({1.5 + 2*sin(60)},-1.2) [whitepiece] () {};
    
    \end{tikzpicture}
}
\vspace{-2pt}
\caption{Immediate win and loss features for the White player in Yavalath.}
\vspace{-8pt}
\label{Fig:WinLossFeaturesYavalath}
\end{figure}

An example in which we were consistently able to observe this problem is the game of Yavalath \cite{Browne2008Thesis}, in which players win the game by constructing lines of four pieces of their colour, but immediately lose if they first construct a line of three pieces of their colour. \reffigure{Fig:WinLossFeaturesYavalath} provides a graphical representation of three features that could be used to detect winning and/or losing moves. The top feature detects winning moves that place a piece to complete a line of four, and the bottom two features detect losing moves that place pieces to complete lines of three. Note that the features that detect losing moves can be viewed as more ``general'' features, in the sense that they will also always be active in situations where the win-detecting feature is active. 

When the set of features is automatically grown over time during self-play, and more ``specific'' features are constructed by combining multiple more ``general'' features \cite{Soemers2019BiasingMCTS}, the loss-detecting features are often discovered before the win-detecting features. These features are -- as expected -- quickly associated with negative weights, resulting in low probabilities $\pi(s, a) \approx 0$ of playing actions $a$ in which loss-detecting features are active. When a win-detecting feature is discovered at a later point in time, the loss-detecting features result in low probabilities $\pi(s, a)$ for most situations in which the win-detecting feature also applies, leading to gradients and update steps close to $0$ despite a strong correlation between feature activity and high values (winning games).

\subsection{Exploratory Policy as Baseline} \label{Subsec:ExploratoryBaseline}

In most (sample-based) policy gradient methods \cite{Williams1992REINFORCE, Sutton2000PolicyGradient, Schulman2016GAE}, there is no longer a $\nabla_{\boldsymbol{\theta}} \pi(s, a)$ term in the gradient estimator. Instead of summing over all actions, updates are typically performed for actions $a$ sampled according to $\pi(s, \cdot)$, which leads to a $\nabla_{\boldsymbol{\theta}} \log \pi(s, a)$ term in the gradient estimator. This gradient, when combined with a softmax-based policy $\pi$, no longer leads to the issue described above. However, there is a closely-related issue in that actions $a$ with low probabilities $\pi(\cdot, a)$ are rarely sampled at all; this problem is generally viewed as a lack of exploration. This is commonly addressed by introducing an entropy regularization term in the objective function, which punishes low-entropy policies \cite{Ahmed2019UnderstandingEntropyRegularization}. That solution is not acceptable for our goals, because it forces an element of exploration in the learned policies -- this is precisely the property inherent in the standard cross-entropy-based approach of Expert Iteration that we aim to avoid. Instead, we propose to use the parameters of a more exploratory policy as a baseline, and train offsets from those parameters using our new policy gradient approach.

Consider a softmax-based policy $\pi_{ce}$, parameterised by a vector $\boldsymbol{\theta}_{ce}$, trained to minimise the standard cross-entropy loss normally used in Expert Iteration. For any given state $s$, this loss is given by \refequation{Eq:CrossEntropyLoss}, where $\boldsymbol{\mathcal{M}}_s(s)$ and $\boldsymbol{\pi}_{ce}(s)$, respectively, denote discrete probability distributions (vectors) over all actions in the state $s$.
\begin{equation}
\label{Eq:CrossEntropyLoss}
\mathcal{L}_{ce}(s) = - \transpose{\boldsymbol{\mathcal{M}}_s(s)} \log \boldsymbol{\pi}_{ce}(s)
\end{equation}

Suppose that $\pi_{ce}$ is defined as a softmax over linear functions of state-action features, parameterised by trainable parameters $\boldsymbol{\theta}_{ce}$, as described in the beginning of this section. Then, the gradient of this loss is given by \refequation{Eq:GradientCrossEntropy}:
\begin{equation}
\label{Eq:GradientCrossEntropy}
\nabla_{\boldsymbol{\theta}_{ce}} \mathcal{L}_{ce}(s) = \sum_{a \in \mathcal{A}(s)} \left[ \left( \pi_{ce}(s, a) - \mathcal{M}_s(s, a) \right) \times \phi(s, a) \right]
\end{equation}
Note that, unlike the gradient in \refequation{Eq:SoftmaxGradientSingleParam}, this gradient does not suffer from the problem that the magnitudes of gradient-based updates are close to $0$ when the trainable policy (in this case $\pi_{ce}$) has (incorrectly) converged to parameters that result in near-zero probabilities for certain state-action pairs. In the example situation described above for Yavalath, we indeed find that a policy trained to minimise this cross-entropy loss is capable of learning high weights for win-detecting features quickly after the feature itself is first introduced.

We propose to exploit this advantage of the cross-entropy loss by defining the logits $z(s, a)$ that are plugged into the softmax of a TSPG-based policy $\pi_{tspg}$ (trained to maximise the TSPG objective of \refequation{Eq:NewObjective}) as follows:
\begin{equation}
\label{Eq:BoostedLogits}
z(s, a) = \transpose{\left( \boldsymbol{\theta}_{ce} + \boldsymbol{\theta}_{tspg} \right)} \boldsymbol{\phi}(s, a).
\end{equation}
Here, $\boldsymbol{\theta}_{ce}$ denotes a parameter vector of a policy $\pi_{ce}$ trained to minimise the cross-entropy loss -- a more ``exploratory'' policy which learns to mimic the exploratory behaviour of MCTS. When training the policy $\pi_{tspg}$ to maximise \refequation{Eq:NewObjective}, we freeze $\boldsymbol{\theta}_{ce}$ and only allow the parameters $\boldsymbol{\theta}_{tspg}$ to be adjusted. This leaves all the gradients and estimators in \refsection{Sec:PolicyGradientsMCTSValues} unchanged. The parameters $\boldsymbol{\theta}_{ce}$ can be viewed as a smart ``initialisation'' of parameters, which is dynamic and can change over time due to its own learning process. The parameters $\boldsymbol{\theta}_{tspg}$ can be viewed as ``offsets'', and the sum of parameters $\boldsymbol{\theta}_{ce} + \boldsymbol{\theta}_{tspg}$ are then the parameters that actually optimise the TSPG objective.

\section{Experiments}

This section describes a number of experiments carried out to compare policies trained to minimise the standard cross-entropy loss of \refequation{Eq:CrossEntropyLoss} with policies trained to maximise the TSPG objective of \refequation{Eq:NewObjective}. All experiments are carried out using a variety of deterministic, adversarial, two-player, perfect information board games.

\subsection{Setup}

All policies are trained using self-play Expert Iteration processes \cite{Silver2017AlphaGoZero, Anthony2017ExIt, Silver2018AlphaZero}. The policies are all defined as linear functions of state-action features \cite{Browne2019StrategicFeatures}, transformed into probability distributions using a softmax, as described in \refsection{Sec:LearningOffsets}. The sets of features grow automatically throughout self-play \cite{Soemers2019BiasingMCTS}.

Experience is generated in self-play, where all players are identical MCTS agents. They use the same PUCT strategy as AlphaGo Zero \cite{Silver2017AlphaGoZero} for the selection phase, with an exploration constant of $2.5$, and a policy $\pi_{ce}$ trained to minimise cross-entropy loss providing bias. All value estimates are in the range $[-1, 1]$, where $-1$ corresponds to losses, $0$ to ties, and $1$ to wins. In the selection phase, unvisited actions are not automatically prioritised; they are assigned a value estimate equal to the value estimate of the parent node. We experiment with policies trained on the cross-entropy objective, as well as policies trained on the TSPG objective, for the play-out phase. Every turn, MCTS re-uses the relevant subtree of the complete search tree generated in previous turns, and runs $1600$ additional MCTS iterations ($800$ in Hex on the $11$$\times$$11$ board, due to high computation time). Actions in self-play are selected proportional to the MCTS visit counts (i.e. sampled from the $\mathcal{M}_s$ distributions in root states $s$).

Every training run described in this section consists of $200$ sequential games of self-play. For every state $s$ encountered in self-play, we store a tuple $\langle s, \mathcal{M}_s, \mathcal{Q}_s \rangle$ in an experience buffer, where $\mathcal{M}_s$ denotes the distribution induced by the visit counts of MCTS, and $\mathcal{Q}_s$ denotes a vector of value estimates $\hat{Q}(s, a)$ for all actions $a \in \mathcal{A}(s)$. Note that the choice to store every encountered state, rather than only one state per full game of self-play, may lead to a poor estimate of the desired distribution over states due to high correlations, but is better in terms of sample efficiency. The maximum size of the experience buffer, which operates as a FIFO queue, is $400$.

After every turn in self-play, we run a single mini-batch gradient descent (or ascent) update per vector of parameters that we aim to optimise (first updating any parameters for cross-entropy losses, and then any parameters for the TSPG objective). Gradients are averaged over mini-batches of up to $30$ samples, sampled uniformly at random from the experience buffer. Updates are performed using a centered variant of RMSProp \cite{Graves2013GeneratingSequences}, with a base learning rate of $0.005$, a momentum of $0.9$, a discounting factor of $0.9$, and a constant of $10^{-8}$ added to the denominator for stability. After every full game of self-play, we add a new feature to the set of features \cite{Soemers2019BiasingMCTS}.

All self-play games are automatically terminated after 150 moves. In the play-out phase of MCTS, play-outs are terminated and declared a tie after $200$ moves have been selected according to the play-out policy.

\begin{figure*}[!t]
    \centering
    \includegraphics[width=.9\textwidth]{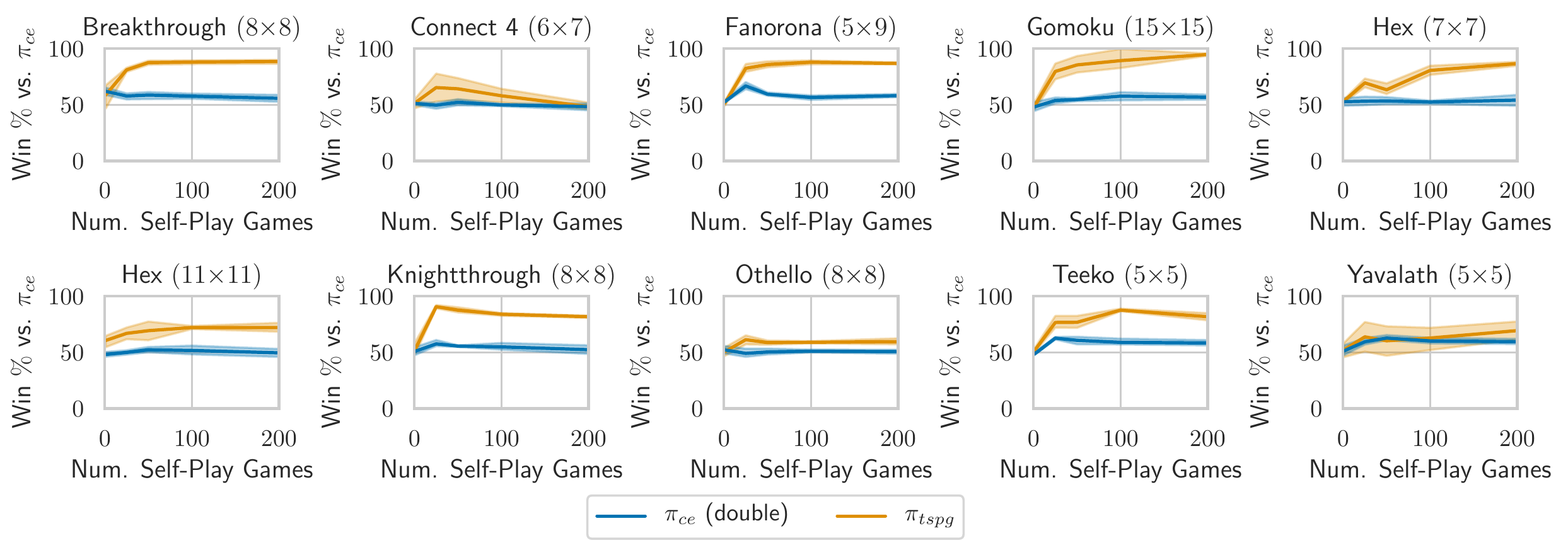}
    \vspace{-10pt}
    \caption{Win percentages of $\pi_{tspg}$ and $\pi_{ce}$ (double) against $\pi_{ce}$, evaluated after 1, 25, 50, 100, and 200 games of self-play.}
    \vspace{-10pt}
    \label{Fig:LearningCurvesSoftmax}
\end{figure*}

Some of the experiments involve evaluating the playing strength of different variants of MCTS after self-play training as described above. We use \textit{Biased MCTS} to refer to a version of MCTS that is identical to the agents used to generate self-play experience as described above, except for that it selects actions to maximise visit count, rather than selecting actions proportional to visit counts, in evaluation games. We use \textit{UCT} to refer to a standard implementation of MCTS \cite{Kocsis2006UCT, Browne2012}, using the UCB1 strategy \cite{Auer2002FiniteTimeMAB} with an exploration constant of $\sqrt{2}$ in the selection phase of MCTS, and selecting actions uniformly at random in the play-out phase. We also allow UCT to reuse search trees from previous turns.

\subsection{Results}

In the first experiment, we compare the raw playing strength of standalone policies trained to either minimise the standard cross-entropy loss, or to maximise the TSPG objective. At various checkpoints during the self-play learning process (after 1, 25, 50, 100, and 200 games of self-play), we run evaluation games between softmax-based policies using the parameters learned at that checkpoint for either objective. We use $\pi_{ce}$ to denote the policy trained on the cross-entropy loss. This is also the same policy that is used throughout self-play to bias the selection phase. We use $\pi_{tspg}$ to denote the policy trained on the TSPG objective. Finally, we use $\pi_{ce}$ (double) to denote a policy that -- like $\pi_{tspg}$ -- uses the parameters of $\pi_{ce}$ as a baseline (see \refsubsection{Subsec:ExploratoryBaseline}), but -- unlike $\pi_{tspg}$ -- again uses the cross-entropy loss to compute offsets from the baseline parameters.

\reffigure{Fig:LearningCurvesSoftmax} depicts learning curves, with the win percentages of $\pi_{tspg}$ and $\pi_{ce}$ (double) against  $\pi_{ce}$ measured at the different checkpoints. We repeat the complete training process from scratch five times with different random seeds, and play 200 evaluation games for each repetition. This leads to five different estimates of each win percentage, each of which is itself measured across 200 evaluation games. We use the sample bootstrap method to estimate $95\%$ confidence intervals \cite{Efron1994IntroductionBootstrap, Henderson2018DeepRLThatMatters} from these five estimates of win percentage per checkpoint, which are depicted as shaded areas.


It is clear from the figure that $\pi_{tspg}$ consistently outperforms $\pi_{ce}$, in many games by a significant margin. We also observe that $\pi_{ce}$ (double) occasionally outperforms $\pi_{ce}$, but generally by a smaller margin than $\pi_{tspg}$.

\reftable{Table:WinPercentages} shows win percentages in evaluation games of a Biased MCTS agent versus UCT. We compare two variants of the Biased MCTS; one where the cross-entropy-based $\pi_{ce}$ (double) policy is used to run MCTS play-outs, and one where the TSPG-based $\pi_{tspg}$ policy is used to run MCTS play-outs. In both cases, we use the final parameters learned after 200 games of self-play. Because our focus in this paper is on evaluating the quality of learned policies or strategies, we run these evaluation games with equal MCTS iteration count limits for all players. Note that this is not representative of playing strength under equal time constraints, since Biased MCTS generally takes more time to run than UCT. However, we do in most games find that Biased MCTS still outperforms UCT under equal time constraints (with most results being slightly improved since our previously-published results \cite{Soemers2019BiasingMCTS}).

Similar to the evaluation in the previous subsection, we include all the different parameters learned from the five different repetitions of training runs in the evaluation. For each vector of parameters resulting from a different repetition, we run 40 evaluation games, for a total of 200 evaluation games across the five repetitions. The different estimates of win percentages from different repetitions are used to construct $95\%$ bootstrap confidence intervals, which are shown in brackets in the table.
In most games, we observe that both variants of Biased MCTS significantly outperform UCT, but play-outs from the cross-entrop-based $\pi_{ce}$ (double) policy often appear to be slightly more informative to the MCTS agent than play-outs based on the TSPG objective.

\begin{table}
\renewcommand{\arraystretch}{1.3}
\caption{Win $\%$ of Biased MCTS vs. UCT (after 200 games of self-play).}
\label{Table:WinPercentages}
\vspace{-3pt}
\centering
\begin{tabular}{@{}lrr@{}}
\toprule
     & \multicolumn{2}{c}{Win $\%$ ($95\%$ bootstrap conf. interval)} \\
     \cmidrule(lr){2-3}
     Game (board size) & $\pi_{ce}$ (double) play-outs & $\pi_{tspg}$ play-outs \\
     \midrule
     Breakthrough $(8$$\times$$8)$ & $100.0$ $(100.0, 100.0)$ & $100.0$ $(100.0, 100.0)$ \\
     Connect 4 $(6$$\times$$7)$ & $76.0$ $(72.0, 80.5)$ & $72.0$ $(67.5, 77.3)$ \\
     Fanorona $(5$$\times$$9)$ & $99.5$ $(99.0, 100.0)$ & $99.2$ $(98.5, 100.0)$ \\
     Gomoku $(15$$\times$$15)$ & $28.0$ $(22.5, 34.0)$ & $18.0$ $(15.5, 20.5)$ \\
     Hex $(7$$\times$$7)$ & $89.5$ $(84.0, 95.0)$ & $88.5$ $(84.0, 93.0)$ \\
     Hex $(11$$\times$$11)$ & $86.5$ $(78.5, 94.5)$ & $71.0$ $(50.0, 95.0)$ \\
     Knightthrough $(8$$\times$$8)$ & $76.5$ $(73.5, 80.0)$ & $63.0$ $(57.5, 69.5)$ \\
     Othello $(8$$\times$$8)$ & $69.0$ $(64.0, 73.5)$ & $69.0$ $(66.0, 72.0)$ \\
     Teeko $(5$$\times$$5)$ & $97.0$ $(94.5, 100.0)$ & $93.8$ $(91.8, 95.0)$ \\
     Yavalath $(5$$\times$$5)$ & $100.0$ $(100.0, 100.0)$ & $98.5$ $(97.5, 99.5)$ \\
\bottomrule
\end{tabular}
\vspace{-10pt}
\end{table}


\begin{figure*}[!t]
    \centering
    \includegraphics[width=.9\textwidth]{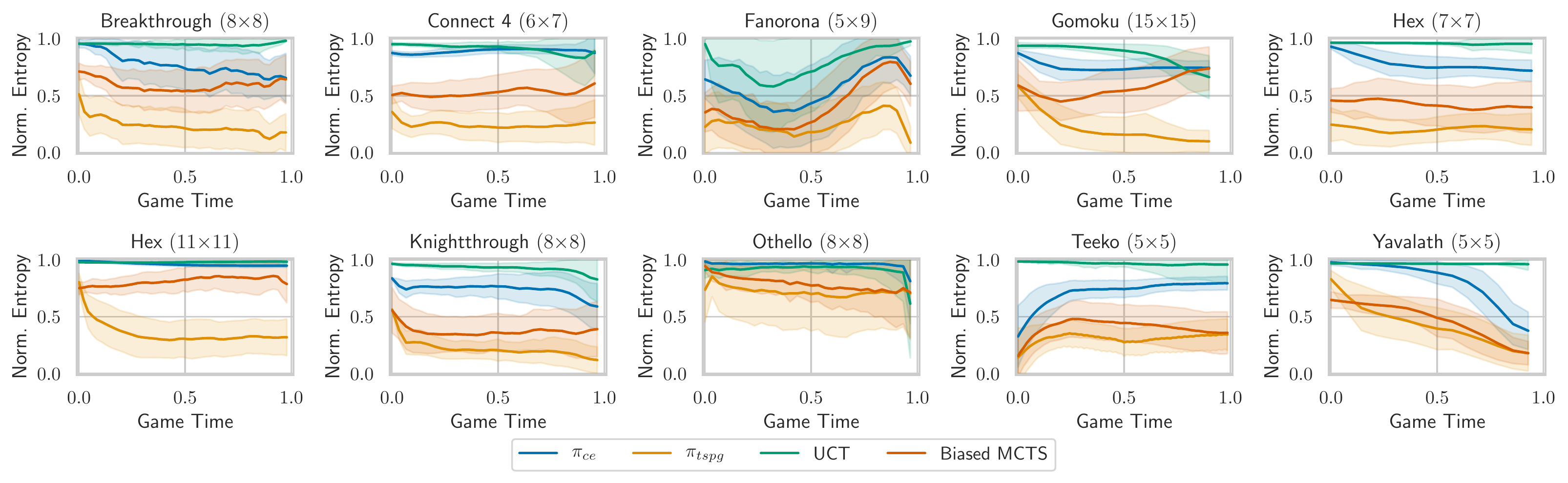}
    \vspace{-6pt}
    \caption{Entropy in distributions over actions for different policies at different stages of a game. Entropy values on the $y$-axis are normalised to adjust for differences in number of legal actions. Game time ($x$-axis) corresponds to turn counter divided by total number of turns played in the corresponding match. For UCT and Biased MCTS, the distributions over actions are derived from the visit counts. Shaded regions depict standard deviation.}
    \vspace{-10pt}
    \label{Fig:EntropyPlots}
\end{figure*}

\begin{figure}[!t]
    \centering
    \includegraphics[width=.8\columnwidth]{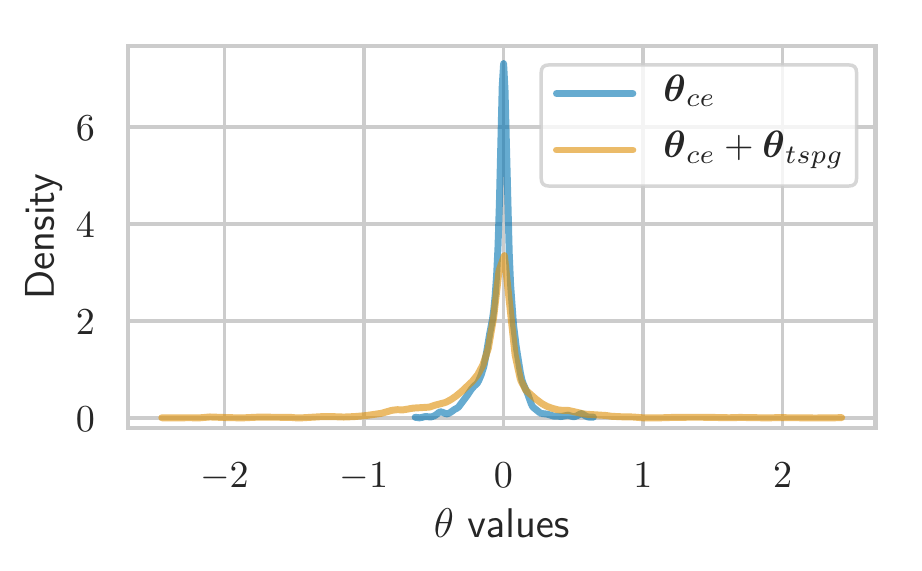}
    \vspace{-10pt}
    \caption{Kernel density estimates for the distributions of $\theta$ values learned when optimising cross-entropy loss ($\boldsymbol{\theta}_{ce}$) or the TSPG objective ($\boldsymbol{\theta}_{ce} + \boldsymbol{\theta}_{tspg}$) in Othello.}
    \label{Fig:KdePlot}
    \vspace{-10pt}
\end{figure}

\reffigure{Fig:EntropyPlots} depicts how the entropy in distributions over actions as computed by a number of different policies varies throughout different stages of the different games. The entropy values are normalised to adjust for differences in the number of legal actions between different games and different stages of the same game. These entropy values were recorded in the evaluation games of Biased MCTS vs. UCT, for which win percentages are shown in \reftable{Table:WinPercentages}.
In most stages of most games, we find that UCT has the highest entropy, followed (often closely) by $\pi_{ce}$, followed by Biased MCTS, finally followed by $\pi_{tspg}$.

\reffigure{Fig:KdePlot} depicts kernel density estimates for the distributions of values in the learned parameter vectors after 200 games of self-play when optimising for the cross-entropy loss ($\boldsymbol{\theta}_{ce}$) or the TSPG objective ($\boldsymbol{\theta}_{ce} + \boldsymbol{\theta}_{tspg}$) in the game of Othello. We observe that the cross-entropy loss leads to a higher peak of parameter values close to $0$, and a shorter range of more extreme parameter values far away from $0$. In all other games (plots omitted to save space), we consistently observed similar differences between the two distributions.

\section{Discussion}

The clear advantage in playing strength that $\pi_{tspg}$ has over $\pi_{ce}$ in \reffigure{Fig:LearningCurvesSoftmax} suggests that the TSPG objective is better suited for learning strong strategies, likely due to the lack of incentive to explore in the objective. The $\pi_{ce}$ (double) policy slightly outperforms $\pi_{ce}$ in some games, which suggests that some small gains in playing strength may simply be due to the increased number of gradient descent update steps that are taken by $\pi_{ce}$ (double) in comparison to $\pi_{ce}$. 

The results in \reftable{Table:WinPercentages} suggest that, despite the higher playing strength of $\pi_{tspg}$, $\pi_{ce}$ (double) may be more informative when used as a play-out policy for MCTS agents. It has previously been observed \cite{Silver2009MCSimulationBalancing, Huang2010MCSimulationBalancingPractice, Graf2016MCSimulationBalancingRevisited} that policies optimised for ``balance'', rather than standalone playing strength, may result in more informative evaluations from MCTS play-outs. Our results suggest that the cross-entropy loss may similarly lead to more balanced policies, leading to a decreased likelihood of biased evaluations.

The entropy plots in \reffigure{Fig:EntropyPlots} show that the distributions over actions recommended by $\pi_{tspg}$ tend to have the lowest entropy, which means that $\pi_{tspg}$ more often approaches deterministic policies, by assigning the majority of the probability mass to only one or a few actions. We expect this to be beneficial for extraction of interpretable strategies from trained policies, because it means that there is more often a clear ranking of actions, and little ambiguity as for which action to pick in any given game state.

An interesting observation is that $\pi_{ce}$ is explicitly optimised (through the cross-entropy loss) for having distributions close to those of Biased MCTS, but it still often has significantly higher entropy than Biased MCTS. In terms of entropy, the distributions resulting from $\pi_{tspg}$ appear to be closer to those of Biased MCTS in many games, despite not being directly optimised for that target.

The results in \reffigure{Fig:KdePlot} suggest that optimising for TSPG rather than cross-entropy loss may make it easier to obtain a clear ranking of features, due to differences between feature weights being more exaggerated, and fewer different features having highly similar weights. We again expect this to be beneficial for interpretation of learned strategies. A comparison to results published on learning balanced play-out policies in Go \cite{Huang2010MCSimulationBalancingPractice} supports the observation described above that the cross-entropy loss may lead to more ``balanced'' \cite{Silver2009MCSimulationBalancing} policies.

\section{Conclusion}

We proposed a novel objective function, referred to as the TSPG objective, for policies in Markov decision processes. Intuitively, a policy that maximises this objective function can be understood as one that selects actions such that, in expectation, an MCTS agent can perform well when playing out the remainder of the episode. We derive a policy gradient expression, which can be estimated using value estimates resulting from MCTS processes. Policies can be trained to optimise this objective using self-play, similar to cross-entropy-based policies in AlphaGo Zero and related research \cite{Silver2017AlphaGoZero, Anthony2017ExIt, Silver2018AlphaZero}. We argue that, due to the lack of a level of exploration in this objective's training target, it is more suitable for goals such as interpretable strategy extraction \cite{Browne2018ModernTechniques, Browne2019StrategicFeatures}.

Across a variety of different board games, we empirically demonstrate that the TSPG objective tends to lead to stronger standalone policies than the cross-entropy loss. Their distributions over actions tend to have significantly lower entropy, which may make it easier to extract clear, unambiguous advice or strategies from them. The TSPG objective also leads to a wider range of different values for feature weights, which can make it easier to separate features from each other based on their perceived importance. 

In future work, we aim to extract interpretable strategies from learned policies, for instance by analysing the contribution \cite{Lundberg2017SHAP} of individual features to the predictions made for specific game positions, or larger sets of positions. The feature representation \cite{Browne2019StrategicFeatures} that we use is generally applicable across many different games, and allows for easy visualisation, which will be beneficial in this regard.

\section*{Acknowledgment.}

This research is part of the European Research Council-funded Digital Ludeme Project (ERC Consolidator Grant \#771292) run by Cameron Browne at Maastricht University's Department of Data Science and Knowledge Engineering. 

\bibliographystyle{IEEEtran}
\bibliography{IEEEabrv,References}

\end{document}